\documentclass[cameraready]{Interspeech}

\title{Ruby-OPD: Enhancing Pronunciation Controllability and Accuracy for Flow-Matching Text-to-Speech via On-Policy Distillation}

\author[affiliation={1}, orcid=0009-0007-3241-6800]{Rui}{Hu}
\author[affiliation={1}, orcid=0009-0006-1490-1610]{Xiaolong}{Lin}
\author[affiliation={1}, orcid=0009-0004-2333-3796]{Zhenpeng}{Zhan}

\address{
    $^1$ Baidu Inc., Shenzhen, China
}

\email{hurui05@baidu.com, linxiaolong01@baidu.com}


\usepackage{comment}
\usepackage{CJKutf8}
\usepackage{multirow}

\begin{document}

\maketitle

\begin{abstract}
    Achieving both high pronunciation accuracy and fine-grained controllability remains a central challenge in Japanese text-to-speech (TTS), as Japanese is an unsegmented language mixing logographic Kanji and phonographic Kana with extensive polyphony. Traditional cascaded systems convert text into phoneme sequences via a front-end, ensuring pronunciation control but discarding lexical semantics and prosodic cues in surface orthography. Conversely, recent zero-shot TTS models directly consume raw orthographic text to improve naturalness, yet suffer from polyphone errors and lack explicit pronunciation control. To bridge this gap, we propose a two-stage framework for flow-matching TTS. First, we investigate Ruby text representations with orthography perturbation to jointly preserve surface forms and target pronunciations for reliable control. Second, we introduce On-Policy Distillation (OPD) on large-scale text-only corpora to enhance pronunciation accuracy on raw orthographic text without paired speech. On the Joyo-Kanji-Yomi benchmark, our model achieves 97.13\% Accuracy, 2.02\% Target PER, 0.70\% Sentence PER, and 4.21\% Sentence CER under raw text while retaining strong controllability.
\end{abstract}

\section{Introduction}

Traditional neural text-to-speech (TTS) systems~\cite{wang17n_interspeech, ren2021fastspeech, pmlr-v139-kim21f} follow a cascaded pipeline comprising a text front-end, an acoustic model, and a vocoder, where the front-end extracts phonemes and prosodic features from raw text. Recently, scaling model capacity and data has driven the rise of zero-shot TTS across both autoregressive (AR)~\cite{du2024cosyvoice, zhou2025voxcpm, hu2026qwen3, liao2026fish, shen2026fireredtts3} and non-autoregressive (NAR)~\cite{le2023voicebox, eskimez2024e2, chen2025f5, zhu2026omnivoice} paradigms. These models bypass complex front-ends and directly consume raw orthographic text, leveraging surface lexical semantics to synthesize natural speech.

In Japanese TTS, however, directly consuming orthographic text poses severe challenges for pronunciation accuracy and controllability. As an unsegmented language, Japanese combines logograms (Kanji) and phonograms (Kana). While Kana maps almost deterministically to pronunciation, Kanji pronunciations vary drastically with morphology and context. For instance, \begin{CJK}{UTF8}{ipxm}秋\end{CJK} (autumn) takes the native pronunciation (\textit{Kun'yomi}) \textit{aki} standalone, the Sino-Japanese pronunciation (\textit{On'yomi}) \textit{sh\={u}} in compounds such as \begin{CJK}{UTF8}{ipxm}晩秋\end{CJK} (late autumn), and the metaphorical pronunciation (\textit{Gikun}) \textit{toki} (\begin{CJK}{UTF8}{ipxm}時\end{CJK}, moment) in classical expressions such as \begin{CJK}{UTF8}{ipxm}此れ誠に危急存亡の秋なり\end{CJK} (this is truly a critical moment of survival).

Early Japanese TTS relied on morphological analyzers with dictionaries~\cite{kudo2005mecab, den2007development, morita2015morphological} and pretrained language models for polyphone disambiguation and accent prediction~\cite{hida2022polyphone, park22b_interspeech, kurihara2024enhancing, hu2026dictionary}. Although phonemic inputs offer direct pronunciation control, stripping away Kanji orthography discards semantic and prosodic cues, limiting naturalness~\cite{yasuda2019investigation}. Conversely, recent end-to-end Japanese TTS models train directly on orthographic text. Irodori-TTS~\cite{irodori-tts} models continuous acoustic latents~\cite{kumar2023high} via conditional flow matching with a Diffusion Transformer (DiT)~\cite{peebles2023scalable}, while Sarashina2.2-TTS~\cite{liu2026sarashina2} uses an AR architecture and mitigates Kanji mispronunciations via local Kana substitution and offline speech synthesis over polyphones. Yet, this still sacrifices orthographic semantics and incurs costly off-policy synthesis.

In this paper, we propose a two-stage framework to enhance pronunciation controllability and accuracy in flow-matching Japanese TTS. First, we introduce Ruby text representations coupled with orthography perturbation to jointly preserve logographic surface forms and target pronunciations for precise pronunciation control. Second, we propose cross-condition On-Policy Distillation (OPD) on large-scale text-only corpora to improve raw-text pronunciation accuracy without paired speech. Experiments on the Joyo-Kanji-Yomi benchmark show that our method substantially improves both pronunciation controllability and accuracy, achieving 97.13\% Accuracy, 2.02\% Target PER, 0.70\% Sentence PER, and 4.21\% Sentence CER.

\section{Method}

\subsection{Ruby text representation}

Ruby is widely used in Japanese books, webpages, and subtitles to annotate the pronunciations of proper nouns, polyphones, rare words, or stylistically specialized terms. Visually, Ruby attaches small Kana characters above or beside target Kanji, simultaneously conveying both orthographic surface form and phonemic pronunciation within a single visual unit.

When linearizing two-dimensional Ruby typography into a sequential text input for a TTS text encoder, we explore two complementary representations:
\begin{itemize}
    \item \textbf{Implicit Ruby}: Target pronunciations are appended directly after annotated Kanji words using parentheses. This aligns naturally with human users' everyday writing and interactive editing habits when manually annotating occasional rare words or proper nouns on the fly;
    \item \textbf{Explicit Ruby}: One limitation of implicit Ruby is the lack of explicit left boundaries for the target span, which can sometimes result in ambiguous annotations in unsegmented text. Explicit Ruby, however, employs explicit boundary markers for both the target span and its pronunciation, making it well suited for unambiguous and precise pronunciation annotation.
\end{itemize}

An example is illustrated in Table~\ref{tab:ruby}. The compound word \begin{CJK}{UTF8}{ipxm}超電磁砲\end{CJK} is canonically pronounced as \begin{CJK}{UTF8}{ipxm}ちょうでんじほう\end{CJK} (\textit{ch\={o}denjih\={o}}). However, in the title \begin{CJK}{UTF8}{ipxm}とある科学の超電磁砲\end{CJK} (\textit{A Certain Scientific Railgun}), it takes the stylized pronunciation \begin{CJK}{UTF8}{ipxm}レールガン\end{CJK} (\textit{r\={e}rugan}). Under implicit Ruby, because this non-standard pronunciation has no morphological correspondence to individual Kanji characters, the left boundary of the annotated span becomes ambiguous, where it could be interpreted as annotating either \begin{CJK}{UTF8}{ipxm}電磁砲\end{CJK} or the entire \begin{CJK}{UTF8}{ipxm}超電磁砲\end{CJK}. In contrast, explicit Ruby eliminates such scope ambiguity by introducing explicit boundary markers.

\begin{table}[th]
  \begin{CJK}{UTF8}{ipxm}
  \eightpt
  \setlength{\tabcolsep}{4pt}
  \caption{An example of orthographic text, implicit Ruby, and explicit Ruby. ｜, 《, and 》 denote the start of the target word, and the start and end of
  the target pronunciation in explicit Ruby, respectively.}
  \label{tab:ruby}
  \centering
  \begin{tabular}{ p{\dimexpr\linewidth-2\tabcolsep\relax} }
    \toprule
    \textit{\textbf{Orthographic text}} \\
    とある科学の超電磁砲 \\
    \midrule
    \textit{\textbf{Implicit Ruby}} \\
    とある科学（かがく）の超電磁砲（レールガン） \\
    \midrule
    \textit{\textbf{Explicit Ruby}} \\
    とある｜科学《かがく》の｜超電磁砲《レールガン》 \\
    \bottomrule
  \end{tabular}
  \end{CJK}
\end{table}

\subsection{Training data construction}

Constructing paired Ruby training data for supervised fine-tuning (SFT) solely via Grapheme-to-Phoneme (G2P) tools is prone to polyphone disambiguation errors that may mismatch the actual pronunciations uttered in the paired speech. To construct precisely aligned Ruby training data, we employ a cross-modal verification approach that leverages information from both text and audio modalities. Specifically, we first use a phoneme recognition model to transcribe the input speech into a pure Kana sequence, capturing the actual pronunciation of the utterance. Then, for the corresponding orthographic text, we use UniDic~\cite{den2007development}, a widely used morphological dictionary, to construct a word lattice encompassing all valid morphological segmentations and candidate pronunciations. Finally, we apply dynamic programming to retrieve the alignment path through the lattice that exactly matches the recognized Kana transcription, precisely pinpointing the character span and true contextual pronunciation of every Kanji word unit in the utterance.

In practical applications, users exhibit diverse annotation habits across different granularities, ranging from short single words to longer multi-word phrases, and may feed either completely unannotated orthographic text or partially annotated Ruby text where only selected words are annotated. To simulate these scenarios, rather than enumerating fixed full-annotation paths after the alignment step, we store the aligned Ruby units and adopt a dynamic multi-granularity sampling strategy during training. First, at the utterance level, we sample from a preset probability distribution to determine whether an utterance is formatted as unannotated, partially annotated, or fully annotated. Specifically, under the partially annotated mode, each Ruby unit is converted into Ruby format with a certain probability and otherwise reverts to raw orthographic text. Furthermore, adjacent Ruby units that are simultaneously selected for annotation are merged with a certain probability into a single contiguous Ruby unit. These data augmentation strategies enable the model to seamlessly handle unannotated raw text, sparse local annotations, and annotations of varying granularities.

\subsection{Orthography perturbation}

In our preliminary experiments, we found that directly training the model with Ruby text failed to make it properly interpret the Ruby format or follow the pronunciation annotations at inference time. Instead, the model often read aloud both the surface form and the annotated pronunciation within Ruby units. We conjecture that because the base model was pretrained on raw orthographic text, it can already infer canonical pronunciations for most Kanji solely from their surface glyphs, leaving little incentive to attend to the pronunciation cues provided in the Ruby annotations.

To address this issue, we keep the target pronunciation annotation intact while stochastically perturbing the surface form of each Ruby unit during training via three complementary operations:
\begin{itemize}
    \item \textbf{Kanji masking}: replaces partial or full Kanji spans in the surface form with special mask tokens. Partial masking encourages the model to align unmasked characters with the target pronunciation to infer span boundaries in implicit Ruby, whereas full masking forces the model to rely on the pronunciation annotation when surface glyph cues are absent;
    \item \textbf{Random substitution}: replaces the surface form in explicit Ruby with random characters or unrelated words of arbitrary length, breaking character-to-mora correspondence and compelling the model to follow the annotated pronunciation regardless of surface glyphs;
    \item \textbf{Semantic substitution}: replaces the surface form in explicit Ruby with semantically similar Kanji words that have distinctly different canonical pronunciations, retrieved via text embeddings over UniDic. This design specifically simulates the aforementioned \textit{Gikun} phenomenon, where the surface form is semantically aligned with both the annotated pronunciation and surrounding context yet phonemically unrelated to its canonical pronunciation. For example, the Kana word \begin{CJK}{UTF8}{ipxm}ワールド\end{CJK} (world), pronounced as \textit{w\={a}rudo}, may be assigned a semantically similar yet orthographically and phonemically distinct surface form \begin{CJK}{UTF8}{ipxm}世界\end{CJK}, canonically pronounced as \textit{sekai}. This strategy ensures that the model strictly adheres to the annotated pronunciation while leveraging surface orthography for semantic and prosodic cues.
\end{itemize}

\subsection{On-policy distillation}

After SFT with Ruby data, the model achieves high pronunciation accuracy when provided with full Ruby annotations. However, as noted earlier, real-world inputs are often unannotated orthographic text or only partially annotated, leaving unannotated spans vulnerable to polyphone disambiguation errors. Attaching an external G2P module at inference time or synthesizing paired text-speech data over polyphones in an off-policy manner~\cite{liu2026sarashina2} are viable workarounds, but they introduce extra pipeline modules or costly offline synthesis steps. Instead, we propose an On-Policy Distillation (OPD) approach for flow-matching TTS that relies solely on paired $(\text{orthographic text}, \text{Ruby text})$ data without requiring parallel audio, improving pronunciation accuracy when Ruby annotations are absent.

In conditional flow matching~\cite{lipman2023flow}, a neural network $v_\theta(x_t, t, c)$ learns a time-dependent velocity field that transports standard Gaussian noise $x_1 \sim \mathcal{N}(0, I)$ at $t=1$ to clean acoustic latents $x_0$ at $t=0$ conditioned on text $c$. With paired speech $x_0$, noisy states are constructed via linear interpolation $x_t = (1-t)x_0 + t x_1$, and the model is trained by minimizing:
\begin{equation*}
    \mathcal{L}_{\text{FM}}(\theta) = \mathbb{E}_{t, x_0, x_1} \Bigl[ \bigl\lVert v_\theta(x_t, t, c) - (x_1 - x_0) \bigr\rVert_2^2 \Bigr].
\end{equation*}
In our OPD framework, the SFT model serves as a frozen teacher $v_\phi$ conditioned on full Ruby text $c_{\text{ruby}}$, while the student $v_\theta$, initialized from $v_\phi$, is conditioned on unannotated orthographic text $c_{\text{raw}}$ (other conditioning is omitted for brevity). Because text-only samples lack ground-truth speech latents $x_0$, we generate the intermediate state $x_t$ online via student rollout~\cite{agarwal2024onpolicy, shaul2026parallel}. Starting from pure Gaussian noise $x_1 \sim \mathcal{N}(0, I)$ with target sequence length predicted by the teacher's duration predictor, we integrate the student's ordinary differential equation (ODE) trajectory up to a randomly sampled timestep $t$ without gradients to obtain the on-policy state $x_t$. At $(x_t, t)$, the student is trained to match the teacher's velocity field:
\begin{equation*}
    \mathcal{L}_{\text{OPD}}(\theta) = \mathbb{E}_{t, x_t} \Bigl[ \bigl\lVert v_\theta(x_t, t, c_{\text{raw}}) - \text{sg}\bigl[v_\phi(x_t, t, c_{\text{ruby}})\bigr] \bigr\rVert_2^2 \Bigr],
\end{equation*}
where $\text{sg}[\cdot]$ denotes the stop-gradient operator, alongside a Huber loss aligning the student's duration predictor with the teacher's. Since the rollout starts from pure noise, $x_t$ contains zero phonemic leakage from ground-truth audio, forcing the student to disambiguate polyphones from context to match the teacher's trajectory. During training, we also interleave the paired speech data used in the SFT stage to anchor acoustic naturalness and retain Ruby controllability.

\section{Experimental setup}

\subsection{Data}

For the SFT stage, we use the Japanese subset of the open-source multilingual speech dataset Emilia~\cite{he2024emilia}. To ensure strict consistency between pronunciation annotations and actual speech content, we transcribe all utterances using kana-whisper\footnote{\url{https://huggingface.co/sbintuitions/kana-whisper}} and filter for utterances whose Kana transcriptions exactly match the word lattice constructed via UniDic~\cite{den2007development}. Combined with duration filtering ($0.5 \sim 30\text{ s}$), we obtain approximately $774\text{ hours}$ of high-quality aligned data.

For subsequent text-only OPD, we follow~\cite{hu2026dictionary} by using entries from UniDic~\cite{den2007development} as seed words and leveraging Gemini 3.1 Pro to synthesize text-pronunciation pairs. For each monophonic and polyphonic entry, we synthesize 5 and 10 contextual sentences, respectively. Specifically, rather than asking the LLM to directly output Ruby annotations, we prompt it to generate sentence-level Kana pronunciation sequences and apply the same lattice alignment method used for Emilia. This yields approximately $680\text{k}$ monophonic sentences and $100\text{k}$ polyphonic sentences.

\subsection{Implementation details}

We build upon the open-source Japanese flow-matching zero-shot TTS model Irodori-TTS-v4.1-Small~\cite{irodori-tts}\footnote{\url{https://huggingface.co/Aratako/Irodori-TTS-v4.1-Small}} as our base model. It initializes its text encoder with a pretrained bidirectional masked language model~\cite{warner2024smarter} and predicts the conditional velocity field of $25\text{ Hz}$, $32$-dimensional continuous DACVAE latents~\cite{kumar2023high} via a DiT backbone.

In the SFT stage, utterance-level annotation modes are sampled with probabilities of $50\%$ full Ruby, $30\%$ partial Ruby, in which each Ruby unit is annotated with a probability of $0.5$, and $20\%$ unannotated raw text. Each annotated Ruby unit is formatted as implicit or explicit Ruby with an equal probability of $0.5$, and adjacent annotated Ruby units are merged with a probability of $0.5$, up to $3$ merges and $8$ characters. For orthography perturbation, we use Qwen3-Embedding-0.6B\footnote{\url{https://huggingface.co/Qwen/Qwen3-Embedding-0.6B}} to retrieve semantic substitution candidates over UniDic~\cite{den2007development}, and each annotated surface span undergoes full-span Kanji masking with a probability of $0.2$, partial character-level masking with a probability of $0.25$, semantic substitution with a probability of $0.2$, and random substitution with a probability of $0.15$. The model is trained for $50\text{k}$ steps with a global batch size of $256$ using AdamW, with a learning rate of $1\times 10^{-4}$ for the main model and $1\times 10^{-5}$ for the pretrained text encoder.

In the OPD stage, we freeze the SFT model as the teacher, which receives $100\%$ full Ruby text, and train the student for $50\text{k}$ steps with a global batch size of $256$, a learning rate of $5\times 10^{-5}$, a text encoder learning rate of $2.5\times 10^{-5}$, and $8$ rollout steps. Each training step samples text-only data with an $80\%$ probability, equally weighted between monophonic and polyphonic subsets with $100\%$ raw text input for the student, and paired speech data with a $20\%$ probability, where student inputs are sampled using $80\%$ raw text, $10\%$ partial Ruby, and $10\%$ full Ruby with the same orthography perturbation settings as in the SFT stage.

\begin{table*}[t]
  \eightpt
  \setlength{\tabcolsep}{4pt}
  \caption{Comparison of pronunciation accuracy, speech quality, and voice cloning performance on the Joyo-Kanji-Yomi benchmark (JKYB) using JVS reference audio prompts. All models directly consume unannotated raw orthographic text at inference time.}
  \label{tab:results_main}
  \centering
  \begin{tabular}{l c c c c c c c c}
    \toprule
    \multicolumn{1}{c}{\multirow{2}{*}[-0.5ex]{\textbf{Model}}} & \multicolumn{4}{c}{\textbf{Pronunciation Accuracy}} & \multicolumn{2}{c}{\textbf{Speech Quality}} & \multicolumn{2}{c}{\textbf{Voice Cloning}} \\
    \cmidrule(lr){2-5} \cmidrule(lr){6-7} \cmidrule(lr){8-9}
    & \textbf{Accuracy} & \textbf{Target PER} & \textbf{Sentence PER} & \textbf{Sentence CER} & \textbf{UTMOSv2} & \textbf{DNSMOS} & \textbf{CAM++} & \textbf{ERes2Net} \\
    \midrule
    Qwen3-TTS 1.7B Base~\cite{hu2026qwen3} & 72.10\% & 23.68\% & 4.38\% & 6.52\% & \textbf{3.50} & 3.23 & 0.757 & 0.779 \\
    FireRedTTS3~\cite{shen2026fireredtts3} & 73.29\% & 22.35\% & 4.04\% & 6.12\% & 3.36 & 3.26 & \textbf{0.833} & \textbf{0.832} \\
    OmniVoice~\cite{zhu2026omnivoice} & 83.66\% & 13.24\% & 2.91\% & 5.18\% & 3.19 & 3.21 & 0.797 & 0.786 \\
    FishAudio S2 Pro~\cite{liao2026fish} & 89.90\% & 8.27\% & 1.52\% & 4.77\% & 3.30 & 3.31 & 0.735 & 0.726 \\
    Sarashina2.2-TTS~\cite{liu2026sarashina2} & 90.42\% & 7.69\% & 3.15\% & 6.26\% & 3.42 & \textbf{3.33} & 0.819 & 0.825 \\
    \midrule
    Irodori-Base~\cite{irodori-tts} & 94.11\% & 4.68\% & 0.97\% & \textbf{4.04\%} & 3.21 & 3.23 & 0.728 & 0.697 \\
    Irodori-Ruby-SFT & 92.32\% & 6.05\% & 1.32\% & 4.40\% & 3.18 & 3.22 & 0.703 & 0.663 \\
    \textbf{Irodori-Ruby-OPD} & \textbf{97.13\%} & \textbf{2.02\%} & \textbf{0.70\%} & 4.21\% & 3.19 & 3.23 & 0.703 & 0.663 \\
    \bottomrule
  \end{tabular}
\end{table*}

\subsection{Evaluation metrics}

To evaluate pronunciation accuracy and controllability, we conduct evaluations on a revised version of the Joyo-Kanji-Yomi benchmark (JKYB)~\cite{liu2026sarashina2}\footnote{\url{https://github.com/Parakeet-Inc/Joyo-Kanji-Yomi-Benchmark-Parakeet-Edition}}. It comprises $13,536$ sentences covering 2,136 Joyo Kanji (regular-use Kanji) and their 4,512 official pronunciations specified by the Agency for Cultural Affairs of Japan. JKYB provides the following metrics:
\begin{itemize}
    \item \textbf{Accuracy}: Percentage of evaluated Kanji pronunciations that exactly match the ground truth;
    \item \textbf{Target PER}: Kana character error rate (Kana-CER) against the ground truth of the evaluated Kanji. Values are capped to 100\%;
    \item \textbf{Sentence PER}: Kana-CER for the full-sentence pronunciation. Values are capped to 100\%;
    \item \textbf{Sentence CER}: Standard CER between the ASR transcription of the utterance and the source text. Values are capped to 100\%.
\end{itemize}

We also evaluate speech quality using UTMOSv2~\cite{baba2024utmosv2} and DNSMOS~\cite{reddy2021dnsmos}, and voice cloning speaker similarity using CAM++~\cite{wang2023cam} and ERes2Net~\cite{chen2023enhanced}. During inference, reference audio prompts are randomly sampled from the JVS corpus~\cite{takamichi2019jvs}, which consists of 100 speakers with 100 normal-style utterances per speaker.

\section{Experimental results}

\subsection{Pronunciation controllability}

We first investigate the effectiveness of Ruby representations for pronunciation control. Using the reference text and ground-truth pronunciations from JKYB, we construct oracle Ruby inputs at both the target-word and full-sentence levels. 

As shown in Table~\ref{tab:results_controllability}, for Irodori-Base, directly replacing the target Kanji word with Kana (Target Kana) effectively improves target pronunciation accuracy, whereas replacing the entire sentence with Kana (Full Kana) causes a notable performance drop at both the target-word and sentence levels, yielding an even higher Sentence PER than unannotated raw text. This confirms that stripping away Kanji orthography across an unsegmented sentence eliminates crucial lexical boundary and semantic cues. In contrast, by jointly preserving surface Kanji and phonemic annotations, both Irodori-Ruby-SFT and Irodori-Ruby-OPD avoid this degradation: expanding annotations from the target word (Target) to the full sentence (Full) maintains high target accuracy while further reducing Sentence PER.

Moreover, explicit Ruby (Expl.) consistently outperforms implicit Ruby (Impl.) across both SFT and OPD models under both target-word and full-sentence settings. While implicit Ruby already provides strong controllability for convenient user interaction, the explicit boundary markers in explicit Ruby effectively eliminate left-boundary scope ambiguity, particularly for multi-character words with non-compositional pronunciations.

\begin{table}[th]
  \eightpt
  \setlength{\tabcolsep}{4pt}
  \caption{Results of pronunciation controllability for Irodori-Base (\textbf{Base}), Irodori-Ruby-SFT (\textbf{SFT}), and Irodori-Ruby-OPD (\textbf{OPD}). \textbf{Target} and \textbf{Full} indicate target-word or full-sentence Kana replacement or Ruby annotation. \textbf{Impl.} and \textbf{Expl.} denote implicit and explicit Ruby.}
  \label{tab:results_controllability}
  \centering
  \begin{tabular}{ l l c c c }
    \toprule
    \textbf{Model} & \textbf{Input} & \textbf{Accuracy} & \textbf{Target PER} & \textbf{Sentence PER} \\
    \midrule
    Base & Raw text & 94.11\% & 4.68\% & 0.97\% \\
    Base & Target Kana & 97.89\% & 1.37\% & 0.56\% \\
    Base & Full Kana & 95.40\% & 2.89\% & 1.06\% \\
    \midrule
    SFT & Target Impl. & 97.56\% & 1.71\% & 0.75\% \\
    SFT & Target Expl. & 98.12\% & 1.26\% & 0.72\% \\
    SFT & Full Impl. & 98.05\% & 1.33\% & 0.53\% \\
    SFT & Full Expl. & 98.20\% & 1.19\% & 0.49\% \\
    \midrule
    OPD & Target Impl. & 98.09\% & 1.30\% & 0.57\% \\
    OPD & Target Expl. & \textbf{98.33\%} & \textbf{1.11\%} & 0.56\% \\
    OPD & Full Impl. & 98.14\% & 1.23\% & 0.45\% \\
    OPD & Full Expl. & 98.24\% & 1.15\% & \textbf{0.42\%} \\
    \bottomrule
  \end{tabular}

\end{table}

\subsection{Main results}

Table~\ref{tab:results_main} compares our models (Irodori-Ruby-SFT and Irodori-Ruby-OPD) against the base model (Irodori-Base) and public state-of-the-art (SOTA) baselines across pronunciation accuracy, speech quality, and voice cloning. At inference time, all models directly consume raw orthographic text without any Ruby annotations. First, while general multilingual TTS models, such as Qwen3-TTS, FireRedTTS3, OmniVoice, and FishAudio S2 Pro, achieve leading performance in speech quality and voice cloning, they still struggle with Japanese polyphonic Kanji pronunciation, exhibiting relatively low Accuracy ($72.10\%\text{--}89.90\%$) and high Target PER ($8.27\%\text{--}23.68\%$). In contrast, Japanese-specialized models such as Sarashina2.2-TTS and Irodori-Base benefit from large-scale Japanese pretraining and demonstrate a clear advantage in pronunciation accuracy, with Irodori-Base achieving $94.11\%$ Accuracy and the lowest Sentence CER ($4.04\%$).

Compared with Irodori-Base, Irodori-Ruby-SFT, fine-tuned solely on Emilia Ruby data, exhibits a slight drop on raw text ($92.32\%$ Accuracy and $6.05\%$ Target PER) and minor shifts in acoustic and speaker metrics, reflecting the limited scale and modest recording quality of the SFT corpus as well as capacity allocated to learning Ruby conditioning. After large-scale text-only OPD, however, Irodori-Ruby-OPD achieves substantial gains on raw text without additional paired speech, boosting Accuracy to $97.13\%$ ($+4.81\%$ over its teacher and $+3.02\%$ over Irodori-Base), cutting Target PER by more than half vs.\ Irodori-Base ($4.68\% \rightarrow 2.02\%$), and achieving the lowest Sentence PER ($0.70\%$). Furthermore, Irodori-Ruby-OPD consistently surpasses Irodori-Ruby-SFT across all four Ruby control modes (Table~\ref{tab:results_controllability}) while maintaining speech quality and voice cloning on par with its teacher. In particular, under target-only annotation (Target Impl. and Target Expl.), Irodori-Ruby-OPD notably reduces Sentence PER compared with Irodori-Ruby-SFT ($0.56\%\text{--}0.57\%$ vs.\ $0.72\%\text{--}0.75\%$), as the remaining unannotated Kanji words in the utterance simultaneously benefit from the enhanced raw-text pronunciation capability distilled during OPD.

\section{Conclusion}

In this paper, we proposed a two-stage framework for flow-matching Japanese TTS that first introduces implicit and explicit Ruby representations coupled with orthography perturbation during SFT to enable fine-grained pronunciation control while preserving surface Kanji semantics. Building upon the Ruby-capable SFT teacher, we further introduced cross-condition On-Policy Distillation (OPD) on large-scale LLM-synthesized text-pronunciation pairs, transferring context-dependent Kanji pronunciation knowledge into raw-text inference without requiring paired speech. Experimental results on the Joyo-Kanji-Yomi benchmark demonstrate that Irodori-Ruby-OPD achieves state-of-the-art pronunciation accuracy on unannotated orthographic text while providing flexible and reliable Ruby-based pronunciation control.

\bibliographystyle{IEEEtran}
\bibliography{mybib}

\end{document}